\pdfoutput=1

\documentclass[11pt]{article}

\usepackage[preprint]{acl}
\usepackage{times}
\usepackage{latexsym}

\usepackage[T1]{fontenc}

\usepackage[utf8]{inputenc}

\usepackage{microtype}

\usepackage{inconsolata}

\usepackage{graphicx}

\usepackage{microtype}
\usepackage{hyperref}
\usepackage{url}
\usepackage{booktabs}
\definecolor{darkblue}{rgb}{0, 0, 0.5}
\hypersetup{colorlinks=true, citecolor=darkblue, linkcolor=darkblue, urlcolor=darkblue}
\usepackage{graphicx} 
\usepackage{tcolorbox}
\usepackage{colortbl,booktabs}
\usepackage{enumitem}
\usepackage{tikz}
\usepackage{multirow}
\usepackage{color}
\usepackage{xcolor}
\usepackage{colortbl,booktabs}
\usepackage{graphicx,wrapfig,lipsum}
\usepackage{fontawesome}
\usepackage[ruled,vlined]{algorithm2e}

\newcommand{\Set}{\mathcal}
\newcommand{\Mat}{\boldsymbol}
\usepackage{amsmath}
\usepackage{amssymb}
\usepackage{mathtools}
\usepackage{amsthm}
\usepackage{tikz}
\newcommand*\circled[1]{\tikz[baseline=(char.base)]{
            \node[shape=circle,draw,inner sep=0.5pt] (char) {#1};}}
\usepackage[capitalize,noabbrev]{cleveref}

\title{All Against Some: Efficient Integration of Large Language Models \\for Message Passing in Graph Neural Networks}

\author{%
  \small Ajay Jaiswal$^{1,2}$, Nurendra Choudhary$^{1}$, Ravinarayana Adkathimar$^{1}$, Muthu P. Alagappan$^{1}$, Gaurush Hiranandani$^{1}$, \\
  \small \textbf{Ying Ding$^{2}$, Zhangyang Wang$^{2}$, Edward W Huang$^{1}$, Karthik Subbian$^{1}$} \\
  \small $^{1}$ Amazon $^{2}$University of Texas at Austin
}

\begin{document}
\maketitle
\begin{abstract}
Graph Neural Networks (GNNs) have attracted immense attention in the past decade due to their numerous real-world applications built around graph-structured data. On the other hand, Large Language Models (LLMs) with extensive pretrained knowledge and powerful semantic comprehension abilities have recently shown a remarkable ability to benefit applications using vision and text data. In this paper, we investigate how LLMs can be leveraged in a \textit{computationally efficient} fashion to benefit rich graph-structured data, a modality relatively unexplored in LLM literature. Prior works in this area exploit LLMs to augment \textit{every} node features in an ad-hoc fashion (not scalable for large graphs), use natural language to describe the complex structural information of graphs, or perform computationally expensive finetuning of LLMs in conjunction with GNNs. We propose \textbf{E-LLaGNN (Efficient LLMs augmented GNNs)}, a framework with an \underline{on-demand LLM service} that enriches message passing procedure of graph learning by enhancing a limited fraction of nodes from the graph. More specifically, E-LLaGNN relies on sampling \textit{high-quality neighborhoods} using LLMs, followed by \textit{on-demand neighborhood feature enhancement} using diverse prompts from our prompt catalog, and finally \textit{information aggregation} using message passing from conventional GNN architectures. We explore several \textit{heuristics-based active node selection strategies} to limit the computational and memory footprint of LLMs when handling millions of nodes. Through extensive experiments \& ablation  on popular graph benchmarks of varying scales (Cora, PubMed, ArXiv, \& Products), we illustrate the effectiveness of our E-LLaGNN framework and reveal many interesting capabilities such as \textit{improved gradient flow in deep GNNs, LLM-free inference ability} \emph{etc}.
\end{abstract}

\section{Introduction}

Graph neural networks (GNNs) have been a powerhouse for handling graph-structured data, often leveraging message passing (MP) at their
core for aggregating knowledge from neighbors. Many real-world graphs from social networks, e-commerce, knowledge graphs, citation networks, and more have rich textual information associated with their nodes and edges. GNNs \citep{Kipf2017SemiSupervisedCW,defferrard2016convolutional,velivckovic2017graph,You2020L2GCNLA,Gao2018LargeScaleLG,chiang2019cluster, chen2018fastgcn,duan2022a,thekumparampil2018attention} typically encode this rich information using either non-contextualized shallow embeddings (\emph{e.g.}, bag-of-words and word2vec) or contextualized language models (LMs, \emph{e.g.}, BERT \citep{devlin2018bert} and RoBERTa \citep{liu2019roberta}), followed by the message-passing paradigm to encode the structural information.

Although LM-based embeddings \citep{zhu2021textgnn} can improve the performance of GNNs by incorporating world knowledge, the improvements are markedly restricted. This is because the embeddings lean on the initial textual information associated with nodes, which might be \textit{noisy, missing, or insufficient for characterizing individual node properties}. Moreover, the \textit{one-to-one mapping} from fixed node text descriptions to LM embeddings limits their ability to encode diverse and unseen knowledge during message passing.  Recently, Large Language Models (LLMs) such as GPT-3\citep{openai2020gpt3} and LLaMA \citep{touvron2023llama} have presented massive amounts of context-aware knowledge and superior semantic comprehension capabilities, all accessisble via prompting. This has encouraged a deep dive to understand how LLMs can benefit the conventional GNN framework. 

\begin{figure*}
  \centering
  \includegraphics[width=0.99\linewidth]{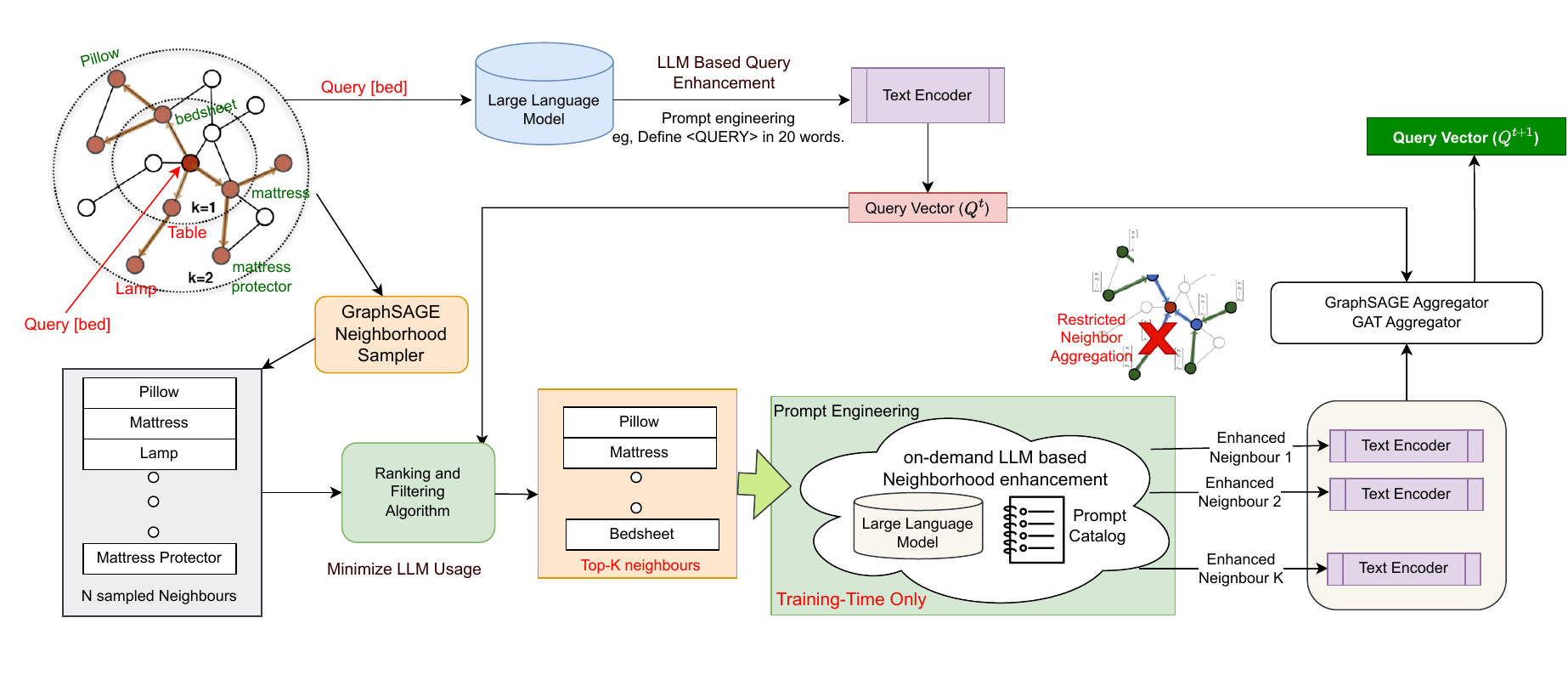}
  \vspace{-2em}
  \caption{Overview of our proposed E-LLaGNN Framework. E-LLaGNN relies on sampling high-quality neighborhoods using LLMs, followed by on-demand neighborhood feature enhancement using diverse prompts from our prompt catalog. Enhanced neighborhood features can be aggregated with the central node using various existing GNN aggregators (\emph{e.g.,} GCN, GraphSAGE, or GAT).}
  \vspace{0.5em}
  \label{fig:main_architecture}
  \vspace{-2em}
\end{figure*}

Recently, several works \citep{Ruosong2023,liuhao2023,JiabinTang2023,Jiayan2023,Xiaoxin2023,Zhikai03393,Jin16595,Zhikai04668,chen2024llaga,Jianxiang09872} have introduced LLMs for graph-structured data, showing surprising performance improvements. Some works use LLMs to extend textual attributes of all the nodes within the graph by harnessing the LLM explanations as features. GNNs are then typically trained on these LLM-tailored features, making model inference dependent on the LLMs to avoid feature distribution shifts. This also makes scaling to large graphs expensive and impractical due to the memory and computational footprint of LLMs. Other groups of works \citep{Ruosong2023} use natural language to describe the geometric structure and node features of the graph. These descriptions are then used to instruction-tune the LLM and perform learning and inference on graphs in a generative manner. Despite this improving the complexity of message passing in GNNs, it is critically important to note the weakness of LLMs in handling long-context information. Recently, \citep{Liang2023} found that LLM performance is often highest when relevant information occurs at the beginning or end of long input contexts and lowest when the information is in the middle. This observation poses a serious question for attempts to use natural language to describe the graph structure, as the order of nodes in the prompt will directly dictate the performance. Lastly, some frameworks fine-tune LLMs to understand graph topology, followed by tightly coupling them with GNNs. This process requires industry-standard hardware for both training and inference. 

Motivated by the aforementioned issues, we ask an important under-explored question in this work: \textit{Does all the nodes within a graph necessitate LLM-based enhancement? Can we efficiently enrich knowledge encoded during GNN training within resource-sensitive settings?} To this end, we propose \textbf{E-LLaGNN}, an efficient and scalable framework that carefully monitors the computational budget of LLM usage while embedding diverse world knowledge in GNNs during training. This computational budget simulates real-world scenarios where allowed LLM queries may be limited. Interestingly, our work finds that E-LLaGNN training with a healthy mixture of LLM-tailored and original node features reduces hard reliance on augmentation at inference and can uniquely facilitate \textbf{LLM-free inference} with marginal compromise in performance which holds significant importance for industrial deployments. 

More specifically, we propose to use LLMs as an on-demand service for high-quality neighborhood selection and augmenting node textual information using diverse (not limited to one-prompt-for-all nodes) prompts from our prompt catalog. This strategy reduces the computational load of prior methods, which have strict requirements for all nodes to be enhanced. Furthermore, having a diverse prompt catalog (unlike a fixed prompt for all) allows rich knowledge to be embedded, since we may use different prompts for the same node if it is selected for augmentation in two different epochs. Note that unlike prior works, the primary motivation of this work is to \underline{\textbf{not}} build SoTA GNN but take the simplest GNN architecture with absolutely no fancy cosmetic modifications like normalizations or skip connections, \emph{etc.} to \textit{capture the pure influence of LLM-enhancement on message passing procedure} of GNN training \footnote{Note that many SoTA architectures are already overfitting to the benchmarks (OGB, Cora, PubMed, \emph{etc.}) and using them directly for E-LLaGNN framework might not truly capture effective benefits of LLM-tailoring during message passing. With complicated backbones, the majority of efforts will boil down to hyperparameter tuning of the model architecture with exorbitant LLM computational costs during training instead of focusing on E-LLaGNN framework.}. However, our proposed framework can be smoothly integrated with any existing GNN architecture to leverage the benefits of LLMs and can be easily scaled to large graphs depending on the node selection techniques. We investigate several heuristics-based active node selection strategies such as degree distribution, PageRank centrality, clustering on original feature space, and original text description length. We experimentally show how they translate to performance, which can guide node selection for augmentation during training. Our contributions can be summarized as:

\begin{itemize}
    \vspace{-0.5em}
    \item We propose the \textbf{E-LLaGNN} framework, which blends LLM capabilities into GNNs as an \textbf{on-demand service} subject to a computational budget during training and can facilitate an \textbf{LLM-free inference}. Our dynamic prompt selection technique enables bringing diverse pretrained knowledge and powerful semantic comprehension abilities, a step ahead of one-for-all prompt setting of prior works.
    \vspace{-0.5em}
    \item We present multiple metrics for active node selection during E-LLaGNN training for on-demand augmentation. Our extensive experiments demonstrate how they translate to performance, which allow LLM integration feasible for industry-scale graphs. Our study reveals that it is \textbf{not} necessary to augment every node to achieve optimal performance. Rather, \textit{tactical augmentation of just a small fraction} of nodes can yield desirable improvements. 
    \vspace{-0.5em}
    \item Our extensive experiments and ablation studies across popular graph benchmarks \texttt{\{Cora, PubMed, OGBN-ArXiv, OGBN-products\}} show how E-LLaGNN can achieve superior performance. Our work also unveils multiple useful insights such as LLM-based enhancement can improve gradient flow across deeper GNN backbone (layers 2, 4, 8), empirical benefits of node-selection proportion, etc.
    
\end{itemize}

\section{Methodology}

\subsection{Preliminaries}
Our work primarily focuses on text-attributed graphs (TAGs), defined as  $G = (\Set{V}, \Set{E}, \Set{T}, \mathbf{A})$ where $\Set{V}$ represents the set of nodes in graph $\Set{G}$ with $\Set{E} \subseteq \Set{V} \times \Set{V}$ edges. Each node $v_i \in \{v_1, v_2, ..., v_n\}$ is paired with textual attributes (\emph{e.g.}, the abstract of papers for citation graphs) $t_i \in  \{t_1, t_2, ..., t_n\}$. The adjacency matrix $\mathbf{A} \in \{0, 1\}^{n \times n}$ represents the graph connectivity, where $\mathbf{A}_{i,j}$ = 1 implies the existence of an edge between nodes $i$ and $j$. For any given node $v_i \in \Set{V}$, we define its immediate neighborhood nodes as $\Set{N}$($v_i$). In our study, we focus on node classification, where given a set of labeled nodes $\mathcal{L} \subset \Set{V}$ and $\mathbf{A}$ with labels $y_{\mathcal{L}}$, our goal is to predict labels for the remaining unlabeled $y_{\Set{V} \setminus \mathcal{L}}$ test nodes.

\subsection{Our Proposed Framework}
\label{sec:proposed-framework}
Our proposed on-demand E-LLaGNN framework is designed with four major components:  \textit{\circled{1} Query Node Enhancement and Encoding, \circled{2} Neighborhood Sampling and Enrichment, \circled{3} On-Demand Neighbourhood Enhancement with Custom Prompt Catalog, and \circled{4} Information Aggregation and Graph Learning}. Given a query node $v_i$, these components help E-LLaGNN to sample high-quality neighbourhoods and ingest LLM expertise within the GNN backbone. E-LLaGNN also explores various active node selection strategies such as PageRank centrality, clustering on the original features, and original text description length, which enable smooth scaling up to large graphs.

Existing pipelines for node feature augmentation are met with challenges such as dependencies on LLMs during inference due to ad-hoc augmentation of entire training dataset leading to distribution shifts in un-augmented test dataset, a need for fine-tuning gigantic LLMs during graph learning, and limited knowledge diversity due to a fixed, single augmentation prompt for all nodes. In contrast, our framework subtly handles these challenges and provides a novel strategy to reap LLM benefits during graph learning. Next, we explain each component of E-LLaGNN in detail.

\paragraph{Query Node Enhancement and Encoding:} At its core, GNN training relies on the message-passing strategy to aggregate knowledge from neighbors. Given the central query node $v_i \in \Set{V}$ with initial textual attribute $t_i \in \Set{T}$, a typical message-passing algorithm aggregates the representative features $x_i$ derived from $t_i$ (\emph{e.g.,} tf-idf or BERT encoding) with the representation features from the neighborhood nodes $u \in$ $\Set{N}$($u$). In our work, we propose to harness the power of LLMs to refine and augment the initial textual representation $t_i \rightarrow t^+_i$ associated with the query node $v_i$. 
The primary motivation is to ingest new LLM knowledge during the query feature aggregation during message passing as well as assist in sampling high-quality and diverse neighborhoods during graph learning. For query enhancement $t_i \rightarrow t^+_i$, we relied on an open-source chatbot trained by fine-tuning LLaMA-1 on user-shared conversations (Vicuna-7B) \citep{vicuna2023}, and LLaMa-2 Chat 7B \citep{touvron2023llama}. We adopted the prompt design from \citep{Xiaoxin2023} 
 for enhancement, and $t^+_i$ is encoded to $x^+_i$ using Sentence-BERT \citep{reimers-2019-sentence-bert}.

\paragraph{Neighborhood Sampling and Enrichment:} In this section, we detail our strategy of sampling high-quality neighbors for aggregation given the central query node $v_i$. We uniformly sample a fixed-size set of neighbors $\Set{N}$($v_i$) from the full neighborhood, which further undergoes filtering as shown in Figure \ref{fig:main_architecture}. We use cosine similarity to re-rank the neighborhood $\Set{N}$($v_i$) using the enhanced query representation $x^+_i$ to create a high-quality fixed batch size of $K$, denoted $\Set{N}_K$($v_i$). This allows us to fix the computational footprint of each batch. It simultaneously minimizes the cost required to use LLMs for neighborhood enhancement while on a set budget. We found that uncontrolled information aggregation from noisy neighbors can hamper the graph learning performance. Therefore, our restricted strategy is not only friendly for computational budgets, but can also filter out noise during the message aggregation. Practically, we found our approach can yield high performance with $K=5$, allowing the per-batch space and time complexities to be fixed.

\paragraph{On-Demand Neighborhood Enhancement with Custom Prompt Catalog:} Given $\Set{N}_K$($v_i$), the goal of this module is to incorporate LLM knowledge into graph learning by enriching the initial text attribute of each node $u \in \Set{N}_K$($v_i$). Our prompt catalog (Table \ref{table:prompt_catalog}) contains prompts designed to provide explanations for key technical concepts/entities, relationships with the central query node, re-writing initial text attributes similar to the query node, elucidating similarities and differences with the central node, etc. Each node $u \in \Set{N}_K$($v_i$) undergoes enhancement ($t_u \rightarrow t^+_u$) with a probability $p$, where $p$ is decided based on the available computational budget. This also ensures that graph learning sees a good mixture of the original text attribute $t_u$ and the LLM-tailored $t^+_u$, which mitigates the dependency on LLM node augmentation during inference. Note that randomly selecting different prompts for the same node across training epochs helps embed diverse knowledge during message passing.

\begin{table}
    \centering
    \resizebox{0.85\columnwidth}{!}{\begin{tabular}{llcc} 
         \toprule
         \multirow{1}{*}{\textbf{Dataset}} &  \multirow{1}{*}{\textbf{Category}} &  \textbf{Raw} & \textbf{Augmented}\\
         \midrule
         Cora & Neural\_Network & 0.3528 & 0.3998\\
         & Probabilistic\_Methods & 0.3891 & 0.4321\\
         & Genetic\_Algorithms & 0.2728 & 0.3028\\
         & Theory  & 0.2951 & 0.3186\\
         & Case\_Based  & 0.4249 & 0.4551 \\
         & Reinforcement\_Learning  & 0.3077 & 0.3530 \\
         & Rule\_Learning  & 0.3646 & 0.3872 \\
         \midrule
         PubMed & Diabetes Type 1  & 0.4674 & 0.5208\\
          & Diabetes Type 2 & 0.4043 & 0.4233\\
          & Diabetes Experimental & 0.4201 & 0.4301\\
         \bottomrule
    \end{tabular}}
    \vspace{-0.5em}
    \caption{Average intra-class cosine similarity among nodes with feature embeddings derived from both original text attributes and LLM-tailored text attributes.}
    \label{table:cosine_sim}
    \vspace{-1.5em}
\end{table}

\subsection{Augmentation and Node Categories}
In this section, we attempt to delve into an interesting question: \textit{How does our textual feature enhancement strategy benefit nodes belonging to different categories?} Table \ref{table:cosine_sim} presents the average intra-class cosine similarity for nodes belonging to different categories. We see that although all categories benefit from our augmentation, some categories (Neural Networks, Probabilistic Methods, and Reinforcement Learning) gain a $\sim5\%$ increase in intra-class similarity. This finding sends a strong signal that in case of limited computational budget, a careful selection of nodes to undergo augmentation during graph learning can still significantly improve the performance. In the next section, we utilize this observation \& explore several node selection strategies to scale E-LLaGNN for large-scale graphs.

\subsection{E-LLaGNN and Large Graphs}
\label{sec:large-graph-node-selection}
The vast knowledge of LLMs ingested during pre-training can significantly benefit graph learning, but using them for node augmentation is computationally expensive. With millions of nodes in large-scale graphs, it is \textbf{impossible} to enhance all nodes to support E-LLaGNN. Therefore, we explore several node selection techniques \footnote{Following \citep{Zhikai04668}, we also attempted to combine PageRank Score and Clustering Distance score using some linear interpolation with coefficient $\alpha$ and found that it to have slightly superior performance ($+0.29\%$ on PageRank). However, this involved fine-tuning $\alpha$. This indicates that instead of relying on a single heuristic, multiple can be integrated to improve the quality of active node selection.} that carefully select a node candidate set to be annotated by LLMs. This is directly applicable to practical use cases where there is a fixed monetary budget for calling LLMs. To this end, we explore the following heuristics for our active node selection during GNN training.

\vspace{-0.5em}
\paragraph{PageRank Centrality:} We used the classical PageRank centrality algorithm from NetworkX \citep{inproceedings}, which estimates which nodes are important in a network based on its topological structure. For every node in our undirected adjacency matrix, we estimate the structural diversity and sorted nodes in ascending order to select the top $k$ nodes to undergo augmentation (K depending on the computational budget).

\vspace{-0.5em}
\paragraph{Clustering Distance:} We employed k-means clustering on the initial features derived from the original text attributes. We set the number of clusters to the number of distinct classes in the graph. We estimate the normalized distance of each node $d_{v_i} \in \Set{V}$ from the cluster centroid, and estimate the density of belonging as 1/(1 + $d_{v_i}$). If distance is low, density is high, and therefore high belonging to the cluster. Depending on the computational budget, we sample nodes with lower density for LLM augmentation.

\vspace{-0.5em}
\paragraph{Text Attribute Length:} One simple heuristic is to examine the initial textual attributes $t_i$ associated with nodes in our graph. We estimate word count in $t_i$ after removing stop words, and sorted nodes by word count. Nodes with the lowest word count are given preference for LLM augmentation during GNN training. Surprisingly, our experiments illustrate that this simple heuristic is highly effective for performance gains. 

\vspace{-0.5em}
\paragraph{Degree Distribution:} Another simple heuristic we used was calculating the degree distribution of each node in the graph, sorting in ascending order and selecting the $k$-th percentile of nodes for augmentation.

 Once nodes are selected, they can be augmented on-demand during E-LLaGNN training, as described in Section \ref{sec:proposed-framework}.

\begin{figure}
    \centering
    \includegraphics[width=\linewidth, trim=0 2em 0 0]{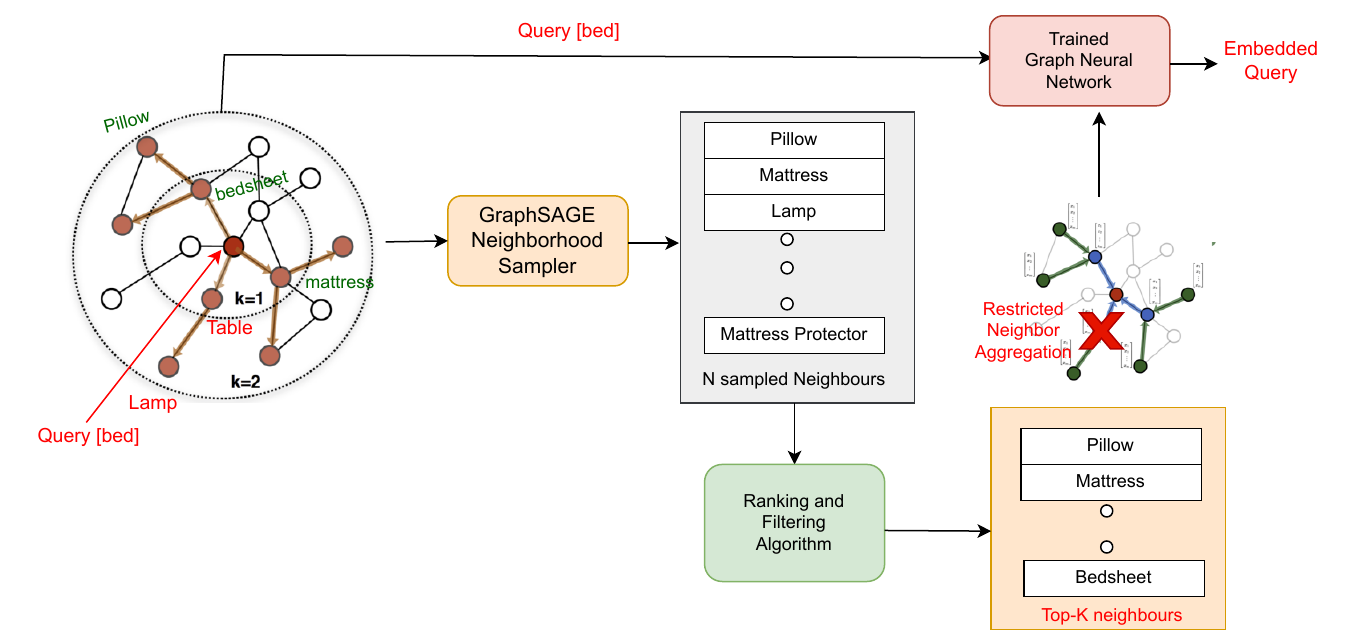}
    \vspace{-0.5em}
    \caption{Overall flow of our LLM-free E-LLaGNN Inference Pipeline which completetly removes LLM dependence during inference. }
    \label{fig:llm-inference}
\end{figure}

\subsection{LLM-free Inference Pipeline of E-LLaGNN}
Despite being a show-stealer due to significant performance benefits, LLM usage faces several headwinds for widespread adaptation because of their enormous computational and memory footprints. One key interesting benefit of E-LLaGNN design is to effectively eliminate LLM usage during inference without significant loss in performance, thereby making it practical for industrial usgage. E-LLaGNN uses LLMs as an \textit{on-demand service} for augmenting node textual information as necessary by randomly selecting prompts from a catalog. This allows GNN training to see a good mixture of both LLM-tailored features as well as the original features. This strategy mitigates over-fitting to LLM-generated textual attributes, thereby improving generalizability even if we perform a completely LLM-free inference. Figure \ref{fig:llm-inference} illustrates the E-LLaGNN inference pipeline, where the high-quality neighborhood $\Set{N}_K$($v_i$) is aggregated with the original features during message passing, followed by the softmax classifier predicting the test labels. In our experiments, we surprisingly found that this E-LLaGNN inference pipeline yields a similar performance to the pipeline that incorporates LLMs during test time.

\begin{table*}
\small
\centering

\vspace{0.5em}
\resizebox{0.95\textwidth}{!}{
\begin{tabular}{p{17cm}}
\toprule
\textbf{Example 1:} A chat between a curious user and an artificial intelligence agent. The assistant gives helpful, detailed, and polite answers to the user's questions. $\backslash$n Provide a detailed explanation for the given title and abstract of a paper in less than 1000 words identifying the key technical concepts and simplifying them for ease of understanding. $\backslash$n $\backslash$n $[$ Start of Title $]$ $\backslash$n \texttt{[content]} $\backslash$n $[$ End of Title $]$. $\backslash$n $[$ Start of Abstract $]$ $\backslash$n \texttt{[content]} $\backslash$n $[$ End of Abstract $]$. \\
\vspace{0.02em}
\textbf{Example 2:} You are a helpful and precise assistant for providing fact-based responses and arguments for user's question. $\backslash$n Rewrite the title and abstract of a paper and provide arguments to support which of the three categories (Diabetes Type 1, Diabetes Type 2, and Diabetes Experimental) is well suited for it to classify. $\backslash$n $\backslash$n $[$ Start of Title $]$ $\backslash$n \texttt{[content]} $\backslash$n $[$ End of Title $]$. $\backslash$n $[$ Start of Abstract $]$ $\backslash$n \texttt{[content]} $\backslash$n $[$ End of Abstract $]$. \\
\vspace{0.02em}
\textbf{Example 3:} A chat between a curious user and an artificial intelligence agent. The assistant gives helpful, detailed, and polite answers to the user's questions. $\backslash$n Summarize the given title and abstract of paper \texttt{P1} and compare its similarities and differences with paper \texttt{P2}. $\backslash$n $\backslash$n $[$ P1: Start of Title $]$ $\backslash$n \texttt{[content]} $\backslash$n $[$ P1: End of Title $]$. $\backslash$n $[$P1: Start of Abstract $]$ $\backslash$n \texttt{[content]} $\backslash$n $[$P1: End of Abstract $]$. $\backslash$n $\backslash$n $[$ P2: Start of Title $]$ $\backslash$n \texttt{[content]} $\backslash$n $[$ P2: End of Title $]$. $\backslash$n $[$P2: Start of Abstract $]$ $\backslash$n \texttt{[content]} $\backslash$n $[$P2: End of Abstract $]$.\\
\bottomrule
\end{tabular}
}
\vspace{-0.5em}
\caption{Examples of prompt curated by us with system message and associated roles USER and ASSISTANT, following \citep{zheng2023judging} to enhance textual attributes of nodes in training dataset.}
\label{table:prompt_catalog}
\vspace{-1.5em}
\end{table*}

\begin{table}
\scriptsize

\centering
\begin{tabular}{lcccccccccccc} 
 \toprule
 \multirow{2}{*}{\textbf{Method}} & \multicolumn{3}{c}{\textbf{Cora}} & \multicolumn{3}{c}{\textbf{PubMed}}\\ 
 \cmidrule(rr){2-4}\cmidrule(rr){5-7}
 
 & 2 & 4 & 8  & 2 & 4 & 8 \\
 \midrule
 GCN & 81.1 & 80.4 & 69.5 & 88.0 & 86.5 & 81.2 \\
 GAT & 81.9 & 80.3 & 31.3 & 88.4 & 87.4 & 79.1 \\
 JKNet  & 79.1 & 79.2 & 75.0  & 87.8 & 88.7 & 77.7 \\
 SGC  & 79.3 & 79.0 & 77.2 & 88.0 & 83.1 & 80.9 \\
 GCNII & 82.2 & 82.6 & 84.2  & 88.2 & 88.8 & 88.3 \\
 IncGCN & 79.2 & 77.6 & 76.5 & 88.5 & 87.7 & 87.9 \\
 
 \midrule
 \rowcolor[gray]{0.9} 
 Ours &  87.6 &  86.5 & 86.1 & 90.3 & 89.3 & 89.4\\
 \rowcolor[gray]{0.9} 
 (std) & $\pm$ 1.2 & $\pm$ 1.7  & $\pm$ 1.3   & $\pm$ 0.1  & $\pm$ 0.1  & $\pm$ 0.4\\
 \bottomrule
\end{tabular}
\vspace{-0.5em}
\caption{Performance comparison (test accuracy \%) of E-LLaGNN framework using underlying 2-, 4-, and 8-layer GNN backbones with respect to SOTA methods on Cora and PubMed. Results reported are averaged across five independent runs.}
\label{table:sota-cora-comparison}
\vspace{-2.0em}
\end{table}

\section{Experiments and Analysis}

\subsection{Dataset and Experimental Setup}
The E-LLaGNN framework is trained and evaluated using four widely recognized graph datasets: ogbn-arxiv~\citep{hu2020open}, ogbn-products~\citep{hu2020open}, PubMed, and Cora~\citep{yang2016revisiting}. We compared the performance benefits for node classification against various SOTA baselines (Table \ref{table:sota-cora-comparison}, \ref{table:sota-ogbn-comparison}). For node attribute enhancement, we relied on an open-source chatbot trained by fine-tuning LLaMA-1 on user-shared conversations (Vicuna-7B) \citep{vicuna2023}. To evaluate our chosen datasets, we closely followed the data split settings and metrics reported by recent benchmarks \citep{duan2022a, chen2021bag}. For neighborhood sampling up to $k$ hops, we adopt the GraphSAGE sampler \citep{hamilton2017inductive}, use cosine similarity for ranking the neighborhood, and perform information aggregation using the MEAN operator. The nodes' original and LLM-tailored text attributes are encoded to embeddings using Sentence-BERT \citep{reimers-2019-sentence-bert}. For the text attribute node selection strategy, we use the NLTK framework for stop word removal and NetworkX for estimating PageRank for nodes in the graph. For additional hyperparameters such as learning rate, weigh decay, and epoch count, we adopted the best settings from \citep{chen2021bag} for Cora and PubMed while relying on \citep{duancomprehensive} for ogbn-arxiv and ogbn-products. For enhancement, we used a temperature of 0.7 and restricted the maximum token length to 2048.

\subsection{Evaluation Protocol}
To validate the effectiveness of E-LLaGNN, we provide experiments to answer several key questions: \circled{1} \textbf{RQ1:} How does on-demand LLM integration help in enriching the knowledge encoded within GNN training, thereby improving its performance? \circled{2} \textbf{RQ2:} How does the amount of nodes selected for augmentation contribute to the performance of GNNs? \circled{3} \textbf{RQ3:} How effective is our proposed framework in facilitating LLM-free inference? \circled{4} \textbf{RQ4:} How do different active node selection techniques help in effectively utilizing LLM benefits to improve GNN performance?

\begin{table*}
\begin{center}
\resizebox{0.75\textwidth}{!}{\begin{tabular}{lcc|cc}
\toprule
\textbf{Method} & \textbf{ogbn-arxiv (Validation)} & \textbf{ogbn-arxiv (Test)} & \textbf{ogbn-products (Validation)} & \textbf{ogbn-products (Test)}\\
\midrule
GraphSAGE & 73.01 $\pm$ 0.89\%  & 71.55 $\pm$ 0.41\% & 81.37 $\pm$ 0.32\% & 80.61 $\pm$ 0.16\%\\
FastGCN &  70.42 $\pm$ 2.11\%     &66.10 $\pm$ 1.06\% & 74.58 $\pm$ 0.97\% & 73.46 $\pm$ 0.20\%\\
LADIES &   65.13 $\pm$ 1.44\%    &62.78 $\pm$ 0.89\% & 75.55 $\pm$ 0.24\% & 75.31 $\pm$ 0.56\%\\
\rowcolor[gray]{0.9} 
\midrule

Ours & 74.32 $\pm$ 0.27\%  & 72.83 $\pm$ 0.74\% & 82.91 $\pm$ 0.56\% & 82.01  $\pm$ 0.33\%\\
\bottomrule
\end{tabular}}
\vspace{-0.5em}
\caption{Performance comparison (test accuracy \%) of the E-LLaGNN framework with popular GNN architectures used for large graphs (ogbn-arxiv and ogbn-products). Results reported are averaged across three independent runs.}
\label{table:sota-ogbn-comparison}
\vspace{-1.2em}
\end{center}
\end{table*}

\begin{table*}
\centering
\resizebox{0.75\textwidth}{!}{\begin{tabular}{llccccccc} 
 \toprule
 \multirow{2}{*}{\textbf{Dataset}} &  \multirow{2}{*}{\textbf{Settings}} & \multicolumn{7}{c}{\textbf{Percentage of Nodes Selected for Augmentation}}  \\ 
 \cmidrule(rr){3-9}
 
 & & 0\% & 5\% &10\% & 25\%  & 50\% & 75\% & 100\%\\
 \midrule
  Cora & Layer 2 & 85.4 $\pm$ 0.7 & 86.1 $\pm$ 1.8 & 87.2 $\pm$ 0.9 & \textcolor{blue}{87.6 $\pm$ 1.2} & 87.1 $\pm$ 0.6 & 86.0 $\pm$ 0.6 & 85.9 $\pm$ 1.3\\
       & Layer 4 & 83.2 $\pm$ 1.3 & 84.3 $\pm$ 1.6 & 84.7 $\pm$ 1.5 & 85.6 $\pm$ 1.0 & \textcolor{blue}{86.5 $\pm$ 1.7} & 85.1 $\pm$ 1.5 & 85.4 $\pm$ 2.3 \\
       & Layer 8 & 82.9 $\pm$ 2.6 & 83.0 $\pm$ 1.9 & 84.1 $\pm$ 2.0 & 85.7 $\pm$ 0.9 & \textcolor{blue}{86.1 $\pm$ 1.3} & 85.9 $\pm$ 1.4 & 85.9 $\pm$ 1.9 \\
    \midrule
  PubMed & Layer 2 & 88.3 $\pm$ 0.7 & 88.4 $\pm$ 0.3 & 89.6 $\pm$ 0.3 & \textcolor{blue}{90.3 $\pm$ 0.1}& 90.2 $\pm$ 0.5 & 89.9 $\pm$ 0.2 & 89.1 $\pm$ 0.1 \\
         & Layer 4 & 87.1 $\pm$ 0.2 & 88.8 $\pm$ 0.1 & \textcolor{blue}{89.3 $\pm$ 0.1} & 89.0 $\pm$ 0.5 & 88.5 $\pm$ 0.2 & 88.7 $\pm$ 0.3 & 88.2 $\pm$ 0.3\\
         & Layer 8 & 86.9 $\pm$ 1.3 & 88.3 $\pm$ 1.1 & 88.5 $\pm$ 1.9 & \textcolor{blue}{89.4 $\pm$ 0.9} & 89.3 $\pm$ 0.7 & 88.8 $\pm$ 1.1 & 87.9 $\pm$ 1.4\\
 \bottomrule
\end{tabular}}
\vspace{-0.5em}
\caption{Performance comparison (test accuracy \%) of the E-LLaGNN framework with underlying 2-, 4-, and 8-layer GNN backbones trained with varying percentages of LLM-tailored nodes from Cora and PubMed. Results reported are averaged across five independent runs.}
\label{table:node-count-cora-comparison}
\vspace{-1.5em}
\end{table*}

\subsection{E-LLaGNN and Popular GNNs}
In this section, we conduct a systematic and extensive study to illustrate how our on-demand LLM-integrated training framework, E-LLaGNN, performs with respect to existing widely adopted GNN architectures. Here, our main focus is to address RQ1 by fairly comparing (similar architecture settings and training hyperparameters) the performance of E-LLaGNN across two small graphs (Cora, PubMed) and two large graphs (ogbn-arxiv and ogbn-products).

Tables \ref{table:sota-cora-comparison} and \ref{table:sota-ogbn-comparison} summarize E-LLaGNN's performance with respect to the performance of several SOTA GNNs following the exact same architecture design. We first observe that across all our candidate datasets, E-LLaGNN significantly outperforms all baselines. More specifically, it achieves improvements in performance over a strong baseline (2-layer GCNII) on Cora by 5.2\% and on PubMed by 2.1\%. In addition, for larger graphs, it surpasses GraphSAGE performance by 1.3\% on ogbn-arxiv and with an impressive gain of 1.6\% on ogbn-products. Moreover, it is important to observe how E-LLaGNN successfully retained performance with increasing depths of the GNN (2$\rightarrow$ 8), significantly outperforming conventional designs like GCN, GAT, \emph{etc.}, and beating robust deep architectures like GCNII. For stability, our results for Cora and PubMed are average scores reported using five independent runs, while for ogbn-arxiv and ogbn-products we used three independent runs. Each run used different seed values selected at random from $[1-100]$ without replacement.

\subsection{Role of Amount of Node Augmentation}
LLMs are expensive and it is impractical to exploit them for enhancing millions of nodes in large real-world graphs. To this end, the E-LLaGNN framework explored active node selection strategies as described in Section \ref{sec:large-graph-node-selection} to predetermine a subset of candidate nodes to undergo augmentation. One natural question arises: \textit{How does the amount of nodes selected impact the performance of E-LLaGNN?}

To this end, Table \ref{table:node-count-cora-comparison} summarizes the performance comparison of E-LLaGNN (with 2, 4, and 8 layers) with varying amount of nodes selected from 0\% to 100\% for Cora and PubMed. Our results illustrate a surprising finding that the best performance is not achieved when all the nodes in the graph undergo augmentation during E-LLaGNN training. We attribute this observation to the E-LLaGNN training overfitting to the LLM-tailored features, which creates distribution shift during the LLM-free inference pipeline. Furthermore, there may be an accumulation of noise throughout node augmentation. This provides a strong positive indication that we should use LLMs judiciously (such as with on-demand LLM augmentation) to reap the maximum benefits while controlling the computational footprint. Note that our findings co-relate with a parallel work \citep{Zhikai04668}, which uses difficulty-aware active selection and filtering strategies to annotate nodes using GPT-3.5/4, then trains GNNs on the high-quality annotated set.

That being said, it is interesting to observe that 100\% augmentation still performs superior to 0\%, which demonstrates the benefits of closely tying LLMs with GNNs. Given our findings, Table \ref{table:node-count-ogbn-comparison} summarizes scaling up E-LLaGNN for large-scale graphs ogbn-arxiv and ogbn-products with millions of nodes. It can be clearly observed that by simply augmenting $\sim$10,000 nodes, we can achieve a noticeable 1.3\% and 1.2\% performance gain on ogbn-arxiv and ogbn-products.

\begin{table}
\tiny
\centering
\begin{tabular}{l|cccccc} 
 \toprule
 \multirow{1}{*}{\textbf{Dataset}}  & \multicolumn{6}{c}{\textbf{Number of Nodes Selected for Augmentation}}  \\ 
 \cmidrule(rr){2-7}
 
 & & 0 & 500 & 1,000  & 5,000 & 10,000 \\
 \midrule
 ogbn-arxiv & & 71.58 & 71.64 & 72.04 & 72.83 & 72.90\\
 ogbn-products & & 80.79 & 80.67 & 81.42 & 81.89 & 82.01  \\
 \bottomrule
\end{tabular}
\vspace{-0.5em}
\caption{Performance comparison (test accuracy \%) of the E-LLaGNN framework trained with varying percentages of LLM-tailored nodes from arxiv \& products.}
\label{table:node-count-ogbn-comparison}
\vspace{-1.2em}
\end{table}

\begin{table}
\tiny
\centering
\begin{tabular}{l|cccc} 
 \toprule
 \multirow{1}{*}{\textbf{Method}}  & \multicolumn{3}{c}{\textbf{Dataset}}  \\ 
 \cmidrule(rr){2-5}
 
 & & Cora & PubMed & ArXiv \\
 \midrule
 PageRank Centrality &  & 87.6 $\pm$ 1.2 & 90.3 $\pm$ 0.1 & 72.9 $\pm$ 0.7 \\
 Clustering Distance & & 86.5 $\pm$ 1.7 & 88.8  $\pm$ 0.8 & 71.7  $\pm$ 1.1\\
 Text Attribute Length & &  87.7 $\pm$ 0.9 &  90.0  $\pm$ 0.6 & 73.4  $\pm$ 0.9\\
 Degree Distribution & & 87.1 $\pm$ 1.2 &  89.4  $\pm$ 0.5 & 72.1   $\pm$ 1.2\\
 \bottomrule
\end{tabular}
\vspace{-0.5em}
\caption{Performance comparison of various active node selection strategies incorporated in the E-LLaGNN framework for selecting nodes for LLM enhancement.}
\label{table:active-node-selection}
\vspace{-2.5em}
\end{table}

\subsection{Performance Comparison of Various Active Node Selection Techniques}
E-LLaGNN for large-scale graphs relies on carefully selecting nodes that maximally benefit the GNN training while staying within the computational budget. To this end, we investigate a series of active node selection techniques: PageRank Centrality, Clustering Distance, Text Attribute Length, and Degree Distribution, as explained in Section \ref{sec:large-graph-node-selection}. Table \ref{table:active-node-selection} illustrates how the aforementioned metrics benefit E-LLaGNN performance on Cora, PubMed, and ogbn-arxiv. Overall, we can observe that among these techniques, surprisingly the simplest average text attribute length performs comparable to PageRank Centrality and outperforms Degree Distribution and Clustering Distance. This highlights an overlooked observation: default features created for nodes with substandard textual descriptions can impair performance, and LLM-based augmentation can significantly benefit the performance of these nodes.

\begin{table}
    \tiny
    \centering
    \begin{tabular}{c|ccc}
    \toprule
    \textbf{Inference Type}& \textbf{Cora} & \textbf{Pubmed} & \textbf{ArXiv}\\
    \midrule
    LLM-free & 87.74 $\pm$ 0.94 & 90.36 $\pm$ 0.15 & 73.43 $\pm$ 0.90 \\
    LLM-incorporated & 87.83 $\pm$ 0.67 & 90.42 $\pm$ 0.07 & 73.79 $\pm$ 0.86\\
    \midrule
    \end{tabular}
    \vspace{-1.0em}
    \caption{Performance of the E-LLaGNN inference pipeline with and without LLM-tailored features.}
    \vspace{-1.5em}
    \label{tab:llm-free-inference}

\vspace{-1.0em}
\end{table}

\subsection{LLM-Free Vs. LLM-Incorporated Inference}
In this section, we attempt to understand how robust E-LLaGNN training is in facilitating an LLM-free inference on the test set. Most approaches \citep{Ruosong2023,Xiaoxin2023,Zhikai03393} that integrate LLMs with GNNs tightly couple them together. By default, this strategy makes the inference too dependent on LLMs, which makes it impractical for industrial deployment due to computational constraints. Table \ref{tab:llm-free-inference} illustrates the performance comparison of E-LLaGNN inference in LLM-free vs. LLM-incorporated settings on Cora, PubMed, ogbn-arxiv. We see that once our framework's GNN backbone is trained using E-LLaGNN's proposed strategy, its performance is impressively similar between the two settings. Across all our candidate datasets, we only observe a marginal performance drop (-0.09\%, -0.06\%, and -0.36\% for Cora, PubMed, and ogbn-arxiv, respectively) for LLM-free inference. These findings can significantly aid the incorporation of LLMs with GNNs for practical purposes.

\subsection{E-LLaGNN Framework \& Gradient Flow}
Despite being powerful tools in learning high-quality node representations, GNNs have been identified as having a limited ability to extract information from high-order neighbors. Furthermore, GNNs suffer from significantly losing their trainability and performance due to poor gradient flow, over-smoothening, information bottlenecks, \emph{etc.} \citep{jaiswalold,wu2019simplifying, Alon2021OnTB}. In this section, we investigate how the E-LLaGNN framework, which incorporates diverse knowledge integration  and careful neighborhood selection during learning, can benefit gradient flow in deep GNNs. Given an 8-layer GCN , we denote the weight parameter for the $l$-th layer $f^{(l)}$ as $\Mat{W}_l$. With node classification cost function $\mathcal{L}$, we calculate the gradient across each layer and effective gradient flow (GF) as:
\begin{align}
&\Mat{g}_1 = \frac{\partial \mathcal{L}}{\Mat{W}_1}, \cdots, \Mat{g}_i = \frac{\partial \mathcal{L}}{\Mat{W}_i}, \cdots, \Mat{g}_L = \frac{\partial \mathcal{L}}{\Mat{W}_L} \\
\label{eqn:gradient_flow} &\textnormal{Gradient Flow:} \operatorname{GF}_{p} = \frac{1}{8} \sum_{n=1}^{8} \|\Mat{g}_n\|_2
\end{align}

\begin{figure}
    \centering
    \includegraphics[width=\linewidth, trim = 0 4em 0 1em]{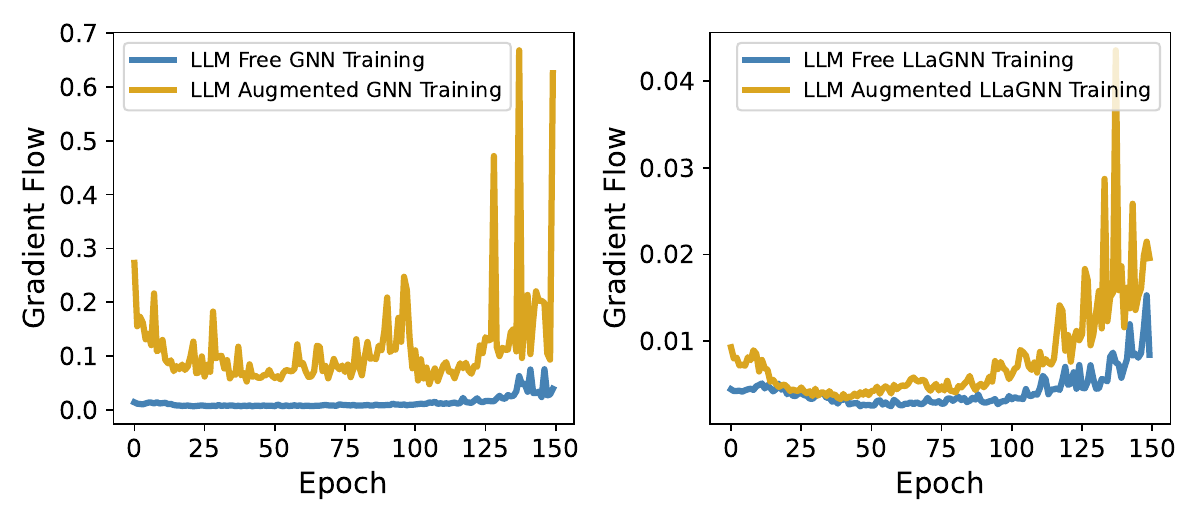}
    \caption{Mean gradient flow across the layers of an 8-layer E-LLaGNN GNN backbone, with and without LLM-based neighborhood enhancement on (a) Cora, and (b) ogbn-arxiv.}
    \label{fig:gradient-analysis}
    \vspace{-1.5em}
\end{figure}
where for every layer $l$, $\frac{\partial \mathcal{L}}{\partial \Mat{W}_{l}}$ represents gradients of the learnable parameters $\Mat{W}_{l}$. As shown in Figure \ref{fig:gradient-analysis}, we provide the gradient flow within the 8-layer LLM-free training vs. LLM-augmented E-LLaGNN training for Cora and ogbn-arxiv. We observe that LLM-augmented E-LLaGNN training significantly improves the gradient flow within the GNN backbone, which is also reflected in the robust performance benefits with deep GNNs (Table \ref{table:node-count-cora-comparison}) across both datasets.

\section{Conclusion}
 We present the E-LLaGNN framework, which blends LLM capabilities into GNNs as an on-demand service, is subjected to a computational budget during training, and facilitates an LLM-free inference. Our study reveals that it is not necessary to augment every node to achieve optimal performance. Rather, tactical augmentation of just a small fraction of nodes can yield desirable improvements. We also present several heuristics-based active node selection strategies such as degree distribution, PageRank centrality, clustering on original feature space, and original text description length, to allow E-LLaGNN to scale to large graphs. Our future work will explore how LLMs can mitigate existing trainability issues such as over-smoothening, information bottlenecks, and heterophily within GNNs.

\section{Limitations}
Our work has several limitations. \underline{\textit{Firstly}}, E-LLaGNN framework is applicable only to text-attributed graphs (TAGs). Like majority of works exploring LLMs integration for graph-structured data, E-LLaGNN relies on initial text-attributes associated with nodes for enrichment using LLMs. \underline{\textit{Secondly,}} Due to cosine similairty as the metric to identify the neighbourhood for augmentation, E-LLaGNN framework is restricted to homophilic datasets. It will be interesting to explore different prioritization/ranking metrics which can can identify heterophilic neighborhood  and observe how LLM enhancement can benefit the challenges of graph learning with heterophilic datasets. \underline{\textit{Thirdly,}} our work primarily focus on the simplest GNN architecture with no cosmetic modification to understand the true impact of LLM enhancement in message passing. We leave further exploration of integrating E-LLaGNN framework with SoTA GNN architecture to observe the additional benefits. \underline{\textit{Fourthly,}} our work exploits LLaMa-7B and Vicuna-7B for feature enhancement. We assume that usage of SoTA high performing  larger set of  LLMs like GPT-4, Claude-3, \emph{etc.} will further improve the performance but left due to large computational and monitory limitations. \underline{\textit{Lastly,}} we do not conduct extensive prompt engineering to identify better prompts for LLMs, which could lead to additional improvements for E-LLaGNN framework. Despite the acknowledged limitations, we hope that our proposed framework and the insights will inspire future work focusing on efficient and compute-constrained integration of LLMs for imporving graph learning.


\bibliography{main}

\clearpage

\appendix

\section{Appendix}
\label{sec:appendix}

\subsection{Background Work}
The recent surge of powerful LLMs has benefitted NLP and vision tasks \citep{ram2023context,li2023cancergpt,guo2022images}, and now they have caught the attention of the graph community. Recent studies \citep{ye2023natural,chen2023label,tang2023graphgpt,guo2023gpt4graph,he2023harnessing,huang2023can} have explored the integration of LLMs with graph-structured data. Some have considered LLMs as enhancers for GNNs, such as extending node features with general knowledge \citep{he2023harnessing, chen2023exploring}. Note that without any careful consideration, these approaches are infeasible to scale to large-scale graphs with millions/billions of nodes. Furthermore, the inference will be dependent on computationally expensive LLMs.

Alternatively, other researchers have attempted to represent graph structures linguistically, then fine-tuning LLMs or employing them to directly address graph tasks \citep{huang2023can,guo2023gpt4graph}. Translating graph structures into natural language is often cumbersome, and the resulting performance is not always optimal. In addition, fine-tuning can compromise the LLM's inherent capabilities in other domains. GraphGPT \citep{tang2023graphgpt} proposes using a pre-trained graph transformer to encode graph structures and align this information with LLM inputs. However, finding a single graph model capable of uniformly encoding structures across various graphs remains challenging. Moreover, recent observations from \citep{Liang2023} found that LLM performance is often highest when relevant information occurs at the beginning or end of the input context, and significantly deteriorates when relevant information is in the middle of long contexts. Therefore, using natural language to describe the graph structure is not straightforward, as the order of nodes in the prompt will directly dictate the performance. Unlike prior works, this paper attempts to explore efficient ways in which LLMs can be incorporated during graph learning through GNNs, while at the same time providing the flexibility to remove LLMs during inference for practical adaptation. Our work makes use of open-source LLMs, unlike the expensive GPT-3.5/4, and found that a prudent selection of a subset of nodes can lead to superior performance rather than augmenting every single node.

\subsection{Additional Related Work}
\subsubsection{Graph Neural Networks}
GNNs have long been at the forefront of graph machine learning. 
They are designed to transform input nodes into compact vector representations, suitable for subsequent classification tasks when paired with a classification head. A common strategy among many GNNs~\citep{kipf2016semi,veličković2018graph,xu2018powerful}, involves a layer-wise message-passing mechanism. This approach enables nodes to progressively aggregate and process information from their immediate neighbors, thereby embedding the nodes into lower-dimensional spaces. Concurrently, a growing body of research~\citep{yun2019graph, ying2021transformers, wu2022nodeformer,chen2022nagphormer}, has been exploring the integration of transformer-based encoders for graph data analysis, opening new avenues for enhancing GNN capabilities.

\subsubsection{Large Language Models}
Enormous increases in scale often permeate systems with unique new behavior. Transformers~\citep{vaswani2017attention}, swiftly after their introduction are scaling every day, dramatically improving the state-of-the-art performance on a wide array of real-world computer vision~\citep{dosovitskiy2020image,Han2020ASO,Touvron2021TrainingDI,Mao2022SingleFA,Zheng2021EndtoEndOD,Parmar2018ImageT}, natural language processing ~\citep{yang2019xlnet,liu2019roberta,talmor2018commonsenseqa,Jaiswal2021RadBERTCLFC,yang2019end,wang2018glue,ding2019cognitive,chowdhery2022palm,wei2022chain,zheng2023outline} applications and leaderboards. Modern LLMs demonstrate significantly augmented capabilities through a substantial increase in the number of parameters. These models can be broadly classified into two categories. The first category encompasses open-source LLMs that users can deploy locally, providing transparent access to model parameters and embeddings \citep{touvron2023llama}. On the other hand, the second category involves LLMs deployed as services, where limitations are imposed on user interfaces. Consequently, users lack direct access to model parameters, embeddings, or logits in this scenario \citep{openai2020gpt3}. We found it interesting that the true potential of open-source LLMs has not be fully explored to assist conventional GNNs and our work is one among few to investigate them. 

\begin{figure}[ht]
    \centering
    \includegraphics[width=\linewidth, trim = 1em 3em 0em 0em]{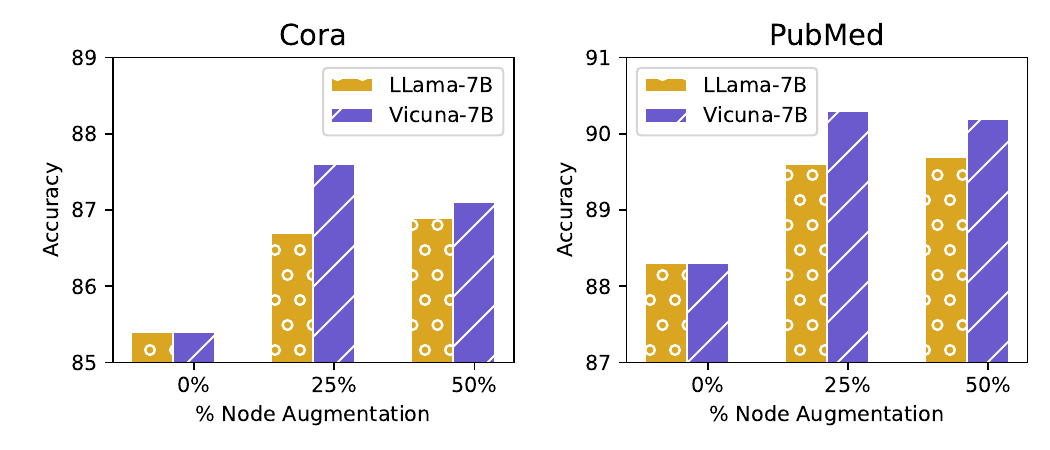}
    \caption{Performance comparison of the 2-layer E-LLaGNN framework trained using varying percentages of node augmentation with LLaMa-7b \& Vicuna-7b on Cora \& PubMed.}
    \label{fig:llm-analysis}
    \vspace{-1.0em}
\end{figure}

\subsubsection{Large Language Models for GNNs}
Recently, Large Language Models (LLMs) have demonstrated extensive context-aware knowledge and advanced semantic comprehension capabilities. This has encouraged a deeper investigation into how LLMs can benefit the conventional Graph Neural Network (GNN) framework. Several studies \citep{ye2023natural,chen2023label,tang2023graphgpt,guo2023gpt4graph,he2023harnessing,huang2023can} have explored the integration of LLMs with graph-structured data. Some have considered LLMs as enhancers for GNNs, using them to augment the textual attributes of graphs, such as extending node features with general knowledge extracted from LLMs \citep{he2023harnessing, chen2023exploring}. However, these approaches still predominantly rely on GNNs for prediction, limiting their generalizability. Alternatively, other researchers have attempted to represent graph structures linguistically, then finetuning LLMs or employing them to directly address graph tasks \citep{huang2023can,guo2023gpt4graph}. Yet, translating graph structures into natural language is often cumbersome, and the resulting performance is not always optimal. While finetuning can enhance performance on graph tasks, it may compromise the LLM's inherent capabilities in other areas. Recently, GraphGPT \citep{tang2023graphgpt} proposes using a pretrained graph transformer to encode graph structures and align this information with LLM inputs.

\subsection{Role of LLMs Selection for E-LLaGNN}
On-demand LLM augmentation forms the core of the E-LLaGNN framework, giving it a unique ability for LLM-free inference. In this section, we explore how the choice of LLM can impact performance. We experimented with LLaMa-7B and Vicuna-7B, and used these pre-trained LLMs to augment various percentages of nodes during the E-LLaGNN training. Figure \ref{fig:llm-analysis} illustrates the performance of the 2-layer E-LLaGNN framework on Cora and PubMed. We observe that irrespective of LLM choice, there is a consistent benefit of carefully embedding LLMs with conventional GNNs. Moreover, our experiments show Vicuna-7B outperforming LLaMa-7B across both datasets.

\begin{table*}
    \centering
    \caption{Graph datasets statistics and download links.}
    \tiny
    \resizebox{0.9\textwidth}{!}{\begin{tabular}{c@{\hspace{1\tabcolsep}}c@{\hspace{1\tabcolsep}}c@{\hspace{1\tabcolsep}}c@{\hspace{1\tabcolsep}}c@{\hspace{1\tabcolsep}}c}
      \toprule
      \textbf{Dataset} & \textbf{Nodes} & \textbf{Edges}  & \textbf{Classes}& \textbf{Download Links}\\
      \midrule
      Cora & 2,708 & 5,429   & 7 & \url{https://github.com/kimiyoung/planetoid/raw/master/data} \\
      \midrule
      Citeseer & 3,327 & 4,732   & 6 & \url{https://github.com/kimiyoung/planetoid/raw/master/data} \\
      \midrule
      PubMed & 19,717 & 44,338   & 3 & \url{https://github.com/kimiyoung/planetoid/raw/master/data} \\
      \midrule
      ogbn-arxiv & 169,343 & 1,166,243   & 40 & \url{https://ogb.stanford.edu/}  \\
      
      \midrule
      ogbn-products &  2,449,029 & 61,859,140  & 47 & \url{https://ogb.stanford.edu/}  \\
       
      \bottomrule
    \end{tabular}}
    
    \label{tab:dataset_details}
\end{table*}

\subsection{Dataset Details}
\label{data}
Table \ref{tab:dataset_details} provided provides the detailed properties and download links for all adopted datasets. We adopt the following benchmark datasets since i) they are widely applied to develop and evaluate GNN models, especially for deep GNNs studied in this paper; ii) they contain diverse graphs from small-scale to large-scale or from homogeneous to heterogeneous; iii) they are collected from different applications including citation network, purchase network, etc.



\section{Code adaptation URL for our baselines}
\label{url_link}
\begin{table*}
    \centering
    \caption{Method and their official implementation used in our work.}
    \tiny
    \resizebox{0.8\textwidth}{!}{\begin{tabular}{l@{\hspace{1\tabcolsep}}l@{\hspace{1\tabcolsep}}}
      \toprule
       Method & Download URL\\
       \midrule
       JKNet\citep{xu2018representation} & \url{https://github.com/mori97/JKNet-dgl}\\
       DAGNN\citep{yang2021dagnn} & \url{https://github.com/vthost/DAGNN}\\
       APPNP\citep{Klicpera2019PredictTP} & \url{https://github.com/gasteigerjo/ppnp}\\
       GCNII\citep{Chen2020SimpleAD} & \url{https://github.com/chennnM/GCNII}\\
       SGC\citep{wu2019simplifying} & \url{https://github.com/Tiiiger/SGC}\\
       ClusterGCN\citep{chiang2019cluster} & \url{https://github.com/benedekrozemberczki/ClusterGCN}\\
       GraphSAINT\citep{zeng2019graphsaint} & \url{https://github.com/GraphSAINT/GraphSAINT}\\
      \bottomrule
    \end{tabular}}
\end{table*}

\end{document}